\documentclass[preprint,tightenlines,superscriptaddress]{revtex4-1}

\pdfoutput=1

\usepackage{graphicx}
\usepackage{amsmath}
\usepackage{amssymb}

\newcommand{\be}{\begin{equation}}
\newcommand{\bea}{\begin{eqnarray}}
\newcommand{\ee}{\end{equation}}
\newcommand{\eea}{\end{eqnarray}}
\def \blank{\mbox{}}

\def\X{\mbox{{\bf X}}}
\def\x{\mbox{{\bf x}}}

\def\Y{\mbox{$\bf{Y}$}}
\def\y{\mbox{$\bf{y}$}}

\def\p{\mbox{$\bf{p}$}}

\def\F{\mbox{$\bf{F}$}}

\def\W{\mbox{{\bf W}}}

\begin{document}
\title{Machine Learning as Statistical Data Assimilation} 

\author{Henry D.~I.\ Abarbanel}
\email{Corresponding author, habarbanel@ucsd.edu}
\affiliation{Marine Physical Laboratory (Scripps Institution of Oceanography), University of California, San Diego, La Jolla, CA 92093-0374, USA}
\affiliation{Department of Physics, University of California, San Diego, La Jolla, CA 92093-0374, USA}
\author{Paul J.\ Rozdeba}
\author{Sasha Shirman}
\affiliation{Department of Physics, University of California, San Diego, La Jolla, CA 92093-0374, USA}

\date{\today}

\begin{abstract}
We identify a strong equivalence between neural network based machine learning (ML) methods and the formulation of statistical data assimilation (DA), known to be a problem in statistical physics. DA, as used widely in physical and biological sciences, systematically transfers information in observations to a model of the processes producing the observations. The correspondence is that layer label in the ML setting is the analog of time in the data assimilation setting. Utilizing aspects of this equivalence we discuss how to establish the global minimum of the cost functions in the ML context, using a variational annealing method from DA. This provides a design method for optimal networks for ML applications and may serve as the basis for understanding the success of ``deep learning.'' Results from an ML example are presented.

When the layer label is taken to be continuous, the Euler-Lagrange equation for the ML optimization problem is an ordinary differential equation, and we see that the problem being solved is a two point boundary value problem. The use of continuous layers is denoted ``deepest learning''. The Hamiltonian version provides a direct rationale for back propagation as a solution method for the canonical momentum; however, it suggests other solution methods are to be preferred.
\end{abstract}

\maketitle


Using enhanced computational capabilities two, seemingly unrelated, ``inverse'' problems have flourished over the past decade. One is machine learning (ML),~\cite{jason1,goodfellow16,deep15} with developments denoted ``deep learning'', in several application areas involving well labeled image recognition~\cite{skin,retina}. The other is data assimilation (DA) in the physical and life sciences. This describes the transfer of information from observations to models of the processes producing those observations~\cite{bennett92,even,abar13}.

This paper demonstrates that these two areas of current investigation are the same at a fundamental level. Each is a statistical physics problem where methods utilized in one may prove valuable for the other. This equivalence provides a unified fashion for the analysis as statistical physics challenges with broad implications for applications across many important areas from medical diagnostics~\cite{skin,retina} to numerical weather prediction~\cite{bennett92,even,abar13}.

We attend to: (1) A variational annealing (VA) method for the action (cost function) for ML and DA that permits the location of the apparent global minimum of that cost function. (2) The notion of analyzing each problem in continuous time or layer, which we call {\tt deepest learning}. It is made clear that one is addressing the same two point boundary value problem~\cite{ye2015physrev,gfomin} with an underlying symplectic structure. Methods abound for solving such two point boundary value problems~\cite{press,hairer06} and for assuring that symplectic structures are respected when time (or layer) is discretized.

The analysis of DA as a statistical physics question~\cite{abar13} is well established, so we begin with multi-layer perceptrons and ML~\cite{jason1,goodfellow16,deep15}. The connection to DA will be made in this context, and an example of the role of statistical physics approaches in ML will be presented.

Briefly recall the formulation of a multi-layer perceptron~\cite{jason1,goodfellow16}. The network comprises an input layer $l_0$, an output layer $l_F$ and ``hidden'' layers $l_1, l_2, ..., l_F - 1$. Each layer has $N$ active units, called ``neurons'', with activity $x_j(l);\;j=1,2,...,N$. This can be generalized to different numbers and different types of neurons in each layer and recurrent connections among neurons at the cost of a notation explosion.

Noisy data is available to layer $l_0$ and layer $l_F$ as $M$ pairs of $L \le N$-dimensional input: $\{y_r^{(k)}(l_0), y_r^{(k)}(l_F)\}$ where $k = 1,2, ..., M$ labels the pairs and $r = 1, 2, 3, ..., L \le N$.

At layers $l_0, l_F$ the network activities are compared to the observations for each input/output pair, and the network performance is assessed using an error metric, often a least squares criterion, or cost function
\be
C_{M}(\x^{(k)}(l),\y^{(k)}(l)) = \frac{1}{M}\sum_{k=1}^{M} \frac{1}{2\,L}\sum_{r = 1}^{L} R_m(r,l)[x^{(k)}_r(l) - y^{(k)}_r(l)]^2.
\label{cost}
\ee
$R_m(r,l)$ is nonzero only at $l = \{l_0,l_F\}$.

The activity of neuron $j$ in layer $l$, $x_j(l)$ is determined by the activity in layer $l-1$, $x_j(l-1)$, via the nonlinear map
\be
x_j^{(k)}(l) = f_j(\x_j^{(k)}(l-1),l) = f_j \biggl(\sum_{i=1}^{N}W_{j,i}(l)x_i^{(k)}(l-1)\biggr).
\label{networkml}
\ee
There are numerous choices how the weight functions act and numerous choices for the nonlinear functions~\cite{jason1,goodfellow16,deep15}.  Minimization of Eq. (\ref{cost}) over all neuron activities $x^{(k)}_{j}(l)$ and weights $W_{j,i}(l)$, {\bf subject to} the model Eq. (\ref{networkml}), is used to determine the weights and the activities in all layers. This constitutes the transfer of information from the input/output pairs to the weights, giving the architecture of the network.

The global minimum over the cost function yields the {\bf path} in an ensemble of networks $\{x^{(k)}_{j}(l), W_{j,i}(l)\}$ which identifies the optimal machine.  Finding the global minimum is an NP-complete problem~\cite{murty87}. This suggests one cannot effectively achieve this unless there is a special circumstance. We have a ``special circumstance''.

The ML problem as described here~\cite{jason1,goodfellow16} assumes there is {\bf no error} in the model Eq. (\ref{networkml}). We relax the equality constraint Eq. (\ref{networkml}) by adding it as a penalty function to the cost function,  defining the ML ``action''
\begin{align}
  A_{ML}(\X) =  \sum_{l=l_0}^{l_F} \left \{C_{M}(\x^{(k)}(l),\y^{(k)}(l)) + \frac{R_f}{2}  \sum_{j=1}^N \left[ x_j^{(k)}(l+1) - f_j \left(\sum_{i=1}^{N}W_{j,i}(l)x_i^{(k)}(l)\right)\right] \right\}^2.
  \label{MLaction}
\end{align}

In the limit $R_f \to \infty$ the equality constraint is restored. Another viewpoint sees the layer-to-layer rule as stochastic with additive Gaussian noise and a diagonal precision matrix $R_f$.

The action is $-\log[P(\X|\Y)]$, with $P(\X|\Y)$ the conditional probability density for the paths of activities and weights $\X = \{\x^{(k)}(l_0), \x^{(k)}(l_1),...,\x^{(k)}(l_F), \W(l)\}$ conditioned on the observations $\Y = \{\y^{(k)}(l_0), \y^{(k)}(l_F)\}$.  The expected value of any function on the path $G(\X)$ is given as
\be
E[G(\X)|\Y] = \left \langle G(\X) \right \rangle  = \frac{\int d\X \,G(\X) \exp[-A_{ML}(\X)]}{\int d\X \exp[-A_{ML}(\X)]}.
\label{expected}
\ee
Interesting choices for $G(\X)$ are $\X$ and moments around this mean path. Evaluating this high dimensional integral gives us the desired weights conditional on the input/output pairs. An effective method for approximating Eq. (\ref{expected}) was introduced by Laplace~\cite{laplace,laplace2}. The minima of $A_{ML}(\X)$ are associated with the maxima of $P(\X|\Y)$ and give the dominant contribution to the integral. If there is an isolated dominant smallest minimum of $A_{ML}(\X)$, it should yield an excellent approximation to Eq. (\ref{expected}).

Now we describe the formulation of a statistical DA problem and this will reveal the equivalence we seek.  In DA one has a physical model represented in discrete time $x_a(t_{k+1}) = f_a(\x(t_k), t_k);\;a=1,2,...,D;\;k=0,1,...,F-1$ of dynamical processes where the state of the model $\x(t_k)$ and physical parameters in the model must be informed by observations $\y(\tau_j)$ of a subset of the dynamical variables of the model. The observations are noisy, and the model has errors; the problem is one of statistical physics.

A general model path integral~\cite{abar13} captures the properties of the conditional probability distribution $P(\X|\Y)$ of the model states and parameters $\X$  conditioned on measurements $\Y=\{\y(\tau_1),\y(\tau_2),...,\y(\tau_F)\}$ within an observation window $[t_0, t_F];\; t_0 \le \tau_k \le t_F$. After this window in which information is transferred to the model, we have an estimate of the full model including unknown parameters and unobserved states. Predictions are made with the completed model and compared to new observations to validate (or not) the model. Validation by prediction is essentially the same as the question of generalization as addressed in machine learning~\cite{jason1,goodfellow16}.

If the observations at the times $\tau_k$ are independent, if the noise in the measurements is Gaussian, with a precision matrix $R_m(r,\tau_k)$, and if the error in the model is taken as additive and Gaussian with precision matrix $R_f(a)$, the DA action, $A_{DA}(\X) = -\log[P(\X|\Y)]$ takes the form
\bea
\sum_{n=1}^{F} \sum_{r=1}^L \frac{R_m(r)}{2} \biggl(x_r(\tau_n) - y_r(\tau_n)\biggr)^2 + \sum_{n=0}^{{\cal N}} \sum_{a=1}^{D}  \frac{R_f(a)}{2} \biggl ( x_a(t_{n+1}) - f_a(\x(t_n),t_n)\biggr)^2.
\label{DAaction}
\eea
Between observations $NI$ steps of the model are taken; ${\cal N} = NI(F+1) - 1$.

Comparing Eq. (\ref{MLaction}) and Eq. (\ref{DAaction}) gives the desired connection between the machine learning formulation (with model error) and the statistical data assimilation formulation: identify layer labels as time $l \Leftrightarrow t$. In ML, measurements provide information at $l_0$ and $l_F$ while in DA, measurements may provide information at many times $\tau_s$ within an observation window $[t_0,t_F]$. Information is received in the ML problem as an ensemble of $M$ pairs rather than as a time series in DA. We call Eq. (\ref{DAaction}) or Eq. (\ref{MLaction}) for DA or ML the standard model as it appears widely in ML and DA studies.

To estimate the expected value integral, we use a numerical optimization method to find the minima of $A_{ML}(\X)$ at fixed $R_m$ and $R_f$.  We have developed a {\it variational annealing} (VA) approach~\cite{ye2014precision,ye2015physrev} to finding the path with the smallest value of the action.  The procedure begins by taking $R_f \to 0$, namely the complete opposite of the value found in usual ML or DA where $R_f \to \infty$ from the outset. In the $R_f = 0$ limit, the action is just a quadratic function of the model variables $\x(l)$ at the times measurements are made, and the minimization is simple: $x_r(l_0) = y_r(l_F)$ and $x_r(l_0) = y_r(l_F)$ for the data presented at the input and output layers. The minimum is degenerate as the weights and hidden layer activities are unknown.  We select these other variables by drawing from a uniform distribution with ranges known from the dynamical range of the state variables. Choose K such initial paths and minimize the action. This gives us an initial collection of K paths $\X^{(0)}$.

\begin{figure}[htbp]
  \centering
  \includegraphics[width=12.0cm]{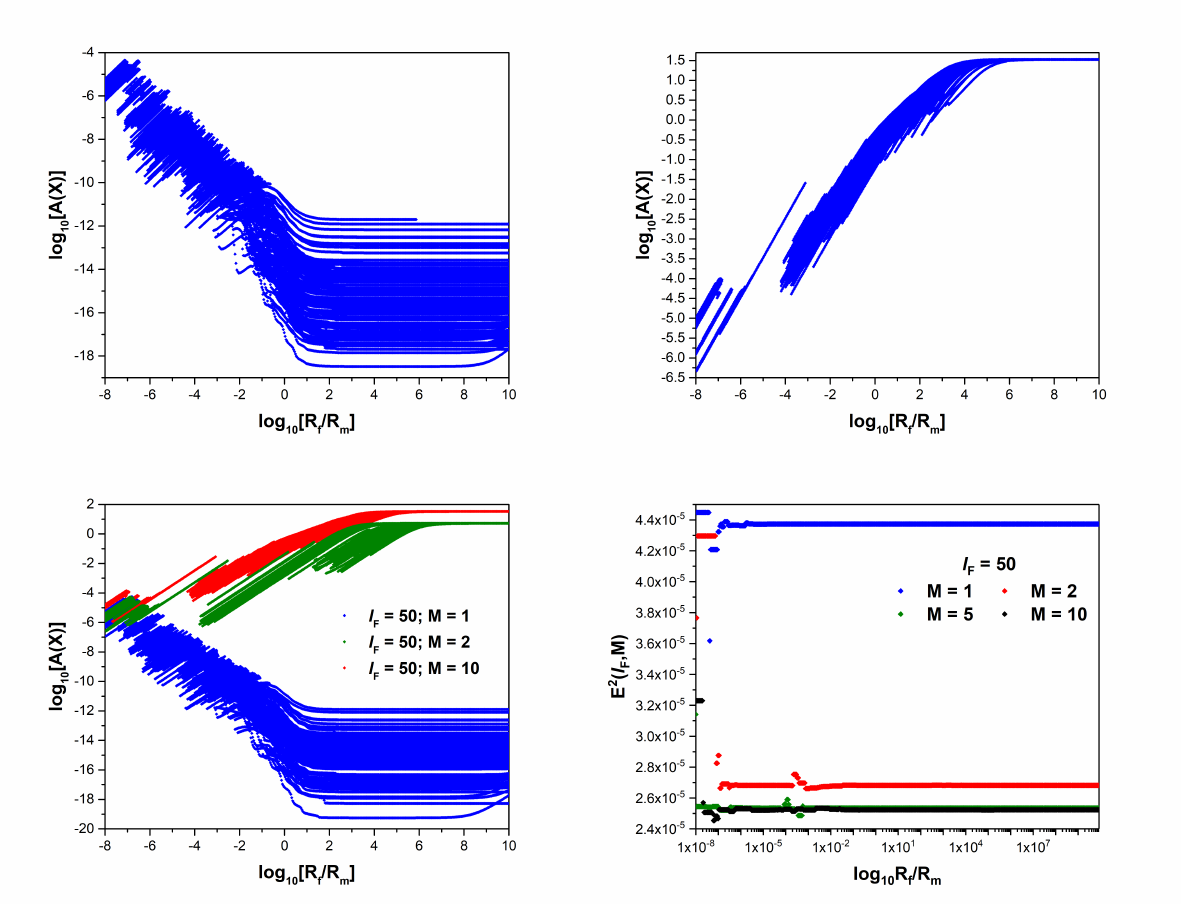}
\caption{Results of using variational annealing to identify smallest minimum action and path of network activities and weights. Using the path giving this minimum action, we use the optimal network to predict on new input/output pairs. Details are in the main body of text.}
\label{prlgraphic}
\end{figure}

Now select a small value for $R_f$, call it $R_{f0}$. With the paths $\X^{(0)}$ as initial choices, perform K optimizations and find K new paths $\X^{(1)}$ for the minimization problem with $R_f = R_{f0}$. This gives us K values of the action $A_0(\X^{(1)})$ associated with the new paths $\X^{(1)}$.  Next increase the value of $R_f$ to $R_f = R_{f0}\alpha$ where $\alpha > 1$. For this new value of $R_f$, we perform the minimization of the action starting with the K initial paths $\X^{(1)}$ from the previous step to arrive at K new paths $\X^{(2)}$. Evaluating the action on these paths $A_0(\X^{(2)})$ now gives us an ordered set of actions that are no longer as degenerate. At the next step increase $R_f$ to $R_{f0}\alpha^2$ and start the minimization of the action with the K paths $\X^{(2)}$ to produce new paths $\X^{(3)}$ and new action levels $A_0(\X^{(3)})$. This procedure is continued until $R_f$ is ``large enough'' which is indicated by at least one of the action levels becoming substantially independent of $R_f$. In Fig. (\ref{prlgraphic}) created from an example ML formulation we will see more clearly what ``large enough'' means in practice.

Effectively VA starts with a problem at $R_f = 0$ where the global minimum is apparent and systematically tracks it and many other paths through slow increases in $R_f$. This is our ``special circumstance.'' In doing the ``tracking'' of the global minimum, one must check that the selected value of $\alpha$ is not too large lest one leave the global minimum and land in another minimum. Checking the result using smaller $\alpha$ is worthwhile.

It is important to note that performing the minimization of $A_0(\X)$ starting with a moderate value of $R_f/R_m \approx 1$ places one in the undesirable situation of the action $A_0(\X)$ having multiple local minima into which any optimization procedure is quite likely to fall, and thus fail to reveal the smallest minimum.

As our goal is to provide accurate estimations of the conditional expected value of functions $G(\X)$ where $\X$, a path in model space, is distributed  as $\exp[-A(\X)]$, we actually do not require convexity of $A(\X)$ as a function in path space. From the point of view of accurately estimating expected values, it is sufficient that the lowest action level be {\bf much} smaller than the second lowest action level.

To illustrate these ideas we created data from a simple ML example of using VA from statistical physics. We prepared a network of the form described with 100 layers and $N = 10$ neurons in each layer. We chose weights for the network, and selected inputs $x^{(k)}_r(l_0);r=1,2,...,N = 10; k=1,2,...$ from a uniform distribution $U[-0.1,0.1]$. These inputs moving through our network produce outputs $x^{(k)}_r(l_F)$. We added Gaussian noise with mean 0 and variance 0.0025 to the inputs and outputs. These make our library of data pairs $\y^{(k)}(l_0), \y^{(k)}(l_F)$.

Our challenge is to choose a model network, knowing only the members of the data pair library, and train it with a subset of these data pairs by minimizing Eq. (\ref{MLaction}) producing a set of weights. We selected $M$ input/output pairs from this library and presented them to a model having $l_F = 10, 20, 50, 100$ layers with 10 inputs $y^{(k)}_i(l_0)$ and 10 outputs $y^{(k)}_i(l_F)$ We investigated $M = 1, 2, 5, 10, 100$. We also examined differing number of inputs and outputs. K = 100 initial conditions were chosen for the numerical optimization procedure. In the numerical optimizations we used the public L-BFGS-B~\cite{byrd,zhu} algorithm.

For each choice of $l_F$ we presented M data pairs and used VA on the action $A_{ML}(\X)$. In Fig. (\ref{prlgraphic}) we present some of the results of these calculations. In the {\bf Upper Left Panel} we show $\log_{10}[A(\X)]$ for $l_F = 20$ M = 1 plotted versus $\log_{10}[R_f/R_m]$ starting with $\log_{10}[R_f/R_m] = -8$ and proceeding to $\log_{10}[R_f/R_m] = 10$ using $\alpha = 1.1$. We see many action levels persist for all values of $\log_{10}[R_f/R_m]$, and observe that the action levels become essentially independent of $\log_{10}[R_f/R_m]$ as $R_f$ increases. Although there is a clear smallest minimum action level, each action level is quite small for this value of M, and none dominate the expected value integral. In the {\bf Upper Right Panel} we now increase M to 10. Now we are providing enough information to the minimization of the action that for large  $\log_{10}[R_f/R_m]$ we see one remaining action, and associated path in activity and weight space. This path should provide accurate estimates of the proposed model. In the {\bf Lower Left Panel} we increase $L_F$ to 50, and show the action levels for $M = 1, 2, 10$ as a function of  $\log_{10}[R_f/R_m]$ . Again for M = 1, which does not provide enough information about the input/output pairs, we see a group of action levels at very small values of the action. When M is increased, one level remains at large  $\log_{10}[R_f/R_m]$.

In the {\bf Lower Right Panel} we examine the prediction error for the model with $l_F = 50$. After training with M = 1,2,5,10 input/output pairs the prediction error is constructed by selecting $k = 1, 2, ..., M_P = 100$ new input/output pairs. We use the model with our estimated weights associated with the path having the global minimum action level to evaluate $\x^{(k)}(l_F)$ from $\y^{(k)}(l_0)$ and compare that with $y^{(k)}_r(l_F)$ from each of the $M_P$ pairs. The square error averaged over $L = 10$ presented components and over $M_P$ pairs $\frac{1}{10\,M_P}\sum_{k=1}^{M_P} \sum_{r=1}^{10} (x^{(k)}_r(l_F) - y^{(k)}_r(l_F))^2$ is displayed. We see that as more information about the pairs in our library is presented to our model (N = 10, L = 10, l$_F$ = 50) the prediction error decreases.

We have used VA to identify the smallest action level and shown that as the number of input/output pairs increases, the prediction error decreases. The prediction capability of the model network also improves when $l_F$ of the model network is increased (not shown here). VA, a tool from the statistical physics of DA has shown its value in designing an ML network.

Analyses of ``deep learning''~\cite{jason1,goodfellow16,deep15} suggest that as the number of layers increases and as the quantity of high quality data increases, predictive capability of the trained network increases. This leads us to examine the continuum limits of the layer label; we call this ``deepest learning''.

The action where the number of layers becomes continuous is  $A(\x(l),\x'(l)) = \int_{l_0}^{l_F} dl\, L(\x(l),\x'(l),l)$ with $\x'(l) = d\x(l)/dl$, and $L(\x(l),\x'(l),l)$ equal to
\be
C_{M}(\x^{(k)}(l),\y^{(k)}(l))  + \sum_{a=1}^N \frac{R_f(a)}{2} \biggl (x'^{(k)}_a(l) - F_a(\x^{(k)}(l),l)\biggr )^2.
\ee
In this $R_m(r,l)$ is nonzero only at $l = \{l_0,l_F\}$ and $\F(\x(l))$ is the vector field for the differential equation in continuous $l$ giving the discrete layer to layer dynamics, Eq. (\ref{networkml}), when layer is made discrete. We call this ``deepest learning''.

The minimization of the action requires that the paths $\x(l)$ in $\{\x(t),\x'(l)\}$ space satisfy the Euler-Lagrange (EL) equation for $\x(l)$
\begin{align}
  x''^{(k)}_a(l) &- \sum_b \Omega_{ab}(\x^{(k)}(l))x'^{(k)}_b(l) \nonumber \\
  &= \frac{\partial \blank}{\partial x^{(k)}_a(l)} \left[ \frac{C_M(\x^{(k)}(l),\y^{(k)}(l))}{R_f}
  + \frac{\F(\x^{(k)}(l),l)^2}{2} \right] + \frac{\partial F_a(\x(l),l))}{\partial l},
\label{euler}
\end{align}
where the skew-symmetric matrix
\begin{align*}
  \Omega_{ab}(\x^{(k)}(l),l) = \partial F_a(\x^{(k)}(l),l)/\partial x_b^{(k)}(l) - \partial F_b(\x^{(k)}(l),l)/\partial x_a^{(k)}(l)
\end{align*}
generates a local rotation. The boundary conditions $\delta x^{(k)}_a(l) p^{(k)}_a(l) = 0;\, l = \{l_0,l_F\}$ 
where $p^{(k)}_a(l) = \partial L(\x(l),\x'(l), l)/\partial x'^{(k)}_a(l)$ is the canonical momentum, must be satisfied by the $\x^{(k)}(l)$. 
In using the Laplace method for Eq. (\ref{expected}) all $\delta \x^{(k)}(l)$ are varied in the minimization of the action, so the appropriate boundary conditions are $\p^{(k)}(l) = 0;\;l = \{l_0,l_F\}$ ~\cite{gfomin,kot,liberzon}.

This shows quite clearly that the minimization problem requires a solution of a two point boundary value problem in $\{\x(l), \x'(l)\}$ space, regardless of whether $l$ is discrete or continuous. If one were to specify $\x(l_0)$, but not $\x(l_F)$, then the boundary conditions for the EL equation require $\p(l_F) = 0$. Examining the Hamiltonian dynamics for this problem then suggest integrating the $\x(l)$ equation forward from $l_0$ and the canonical momentum equation backward from $l_F$. This is back propagation. It answers a different question than evaluating Eq. (\ref{expected}).

The EL equation Eq. (\ref{euler}) is a necessary condition for a minimum of the action, and shows that the dynamics in $l$ involves a rotation generated by $\Omega_{ab}(\x^{(k)}(l))$ forced by the gradient of a generalized ``potential'' involving the measurement error $C_{M}(\x^{(k)}(l),\y^{(k)}(l))$. This suggests an interesting analogy to the motion of a charged particle in a magnetic field in $NM$ dimensions. The potential expansion or contraction of orbits of $\x^{(k)}(l)$ is under control because of the compact structure of the rotations so generated. In the Hamiltonian formulation where backprop is employed, this balancing aspect of the Jacobians $ \partial F_a(\x^{(k)}(l),l)/\partial x_b^{(k)}(l) $ is split between the coordinate equation and the canonical momentuam equation and may lead to unstable or numerically quite difficult issues in its implementation. The solution in Lagrangian coordinates $\{\x(t),\x'(l)\}$ avoids this and retains the symplectic nature of the solutions~\cite{marsdenwest,qjrms17}. It could be that making back propagation explicitly symplectic~\cite{hairer06} could address this issue.

We have shown the equivalence of machine learning problems as posed in a standard manner~\cite{jason1,goodfellow16} and statistical data assimilation problems similarly posed. Both are seen as statistical physics questions where many valuable tools are available for their analysis. Using a variational annealing approach to a simple ML problem, we showed that one may use the hyper-parameter $R_f$ associated with a measure of model error as a design tool to identify the smallest minimum of the action (- $\log$[Conditional Probability]) and thus the dominant contribution to expected value integrals of interesting quantities of statistical models.

While this short note does not discuss recurrent networks~\cite{jordan,elman,parlos} or networks with more structure than multi-layer perceptrons, those can be covered in the framework we have presented.

The detailed results of the way $R_f$ permits design of learning networks provides insight into how ``deep'' a network must be to accurately represent complex information borne in input/output pairs.

The equivalence of the two sets of questions presents opportunities to utilize methods developed in one area, for example meteorology~\cite{even}, in the other, and vice versa. Using the fact that both are statistical physics problems may allow the use of powerful methods not heretofore employed to be used in both areas of research.

By introducing the idea of continuous layers into a ML context (deepest learning) we see that fundamentally one is faced with the solution of two point boundary value problem in ML, and that a Lagrangian approach to the issue of minimizing the action may be more stable and efficient than the Hamiltonian approach wherein back propagation has traditionally been employed.

\section*{Acknowledgments}
Partial support from the MURI Program (N00014-13-1-0205) sponsored by the Office of Naval Research is acknowledged, as is support for S.\ Shirman from the ARCS Foundation.


\end{document}